\documentclass[10pt,twocolumn,letterpaper]{article}

\usepackage{3dv}
\usepackage{times}
\usepackage{epsfig}
\usepackage{graphicx}
\usepackage{amsmath}
\usepackage{amssymb}
\usepackage{url}
\threedvfinalcopy 

\def\threedvPaperID{188} 

\begin{document}

\title{Res3ATN - Deep 3D Residual Attention Network for Hand Gesture Recognition in Videos}

\author{Naina Dhingra,   Andreas Kunz\\
ETH Zurich, Innovation Center Virtual Reality\\
Leonhardstrasse 21, 8092 Zurich, Switzerland\\
{\tt\small (ndhingra, kunz)@iwf.mavt.ethz.ch}
}
\maketitle

\begin{abstract}
Hand gesture recognition is a strenuous task to solve in videos. In this paper, we use a 3D residual attention network which is trained end to end for hand gesture recognition. Based on the stacked multiple attention blocks, we build a 3D network which generates different features at each attention block. Our 3D attention based residual network (Res3ATN) can be built and extended to very deep layers. Using this network, an extensive analysis is performed on other 3D networks based on three publicly available datasets. The Res3ATN network performance is compared to C3D, ResNet-10, and ResNext-101 networks. We study and evaluate our baseline network with different number and position of attention blocks. The comparison shows that the 3D residual attention network with 3 attention blocks is robust in attention learning and can classify the gestures with better accuracy, thus outperforming existing networks. 
   
\end{abstract}

\section{Introduction}

In conversations with other people, we use different types of gestures. In such discussions, non-verbal communication (NVC) is an important part since it could carry up to 55\% of the overall communication \cite{knapp2013nonverbal,frith2009role, gupta2016deep}. In total, 136 different gestures were listed by \cite{brannigan1972human}. Thus, recognizing these gestures - and in particular hand gestures - is crucial to understand the implicit communication in the conversation.

While such NVCs are intuitively understood by human beings, they turn out to be difficult to be recognized and interpreted by machines. The application fields of such automated gesture recognition by machines are manifold, they reach from automated gesture recognition for robots \cite{nickel2007visual}, understanding the psychological factors, up to the evaluation and recognition of the sign language \cite{pigou2014sign, KingkanOH18}. Gesture recognition also plays a huge role in advanced driver assistance systems (ADASs) \cite{baradel2018glimpse}. Here, vision-based hand gesture detection systems are employed for the interaction of the driver with vehicle. They are used for implementing touch-less sensors  \cite{molchanov2015hand}. These sensors help the drivers to interact with secondary functions such as music, heating, etc. which improves the safety and comfort while driving. Furthermore, hand gesture recognition have wide applications in various fields, such as, robotic imitation learning, virtual/augmented reality, tele-operations, security operations, etc.

Hand gestures form a part of our conversation and are important to fully understand the topic being discussed. However, visually impaired people are not able to access these hand gestures and consequently may not be able to easily follow the conversation \cite{PETRA_2019/GuentherKDFHMMK19}. Here, a real-time implementations of the hand gesture detection and the corresponding output to the interface of the blind user could help in addressing this issue.

In this paper, we address the hand gesture detection using a deep learning framework. Motivated by the success of the attention networks and recent advances in deep residual networks, we use a special kind of network, i.e. 3D residual attention network which is an end to end trainable network to classify the hand gestures given in the video frames. Similar to the residual attention network for image classification \cite{wang2017residual}, we use multiple layers of attention blocks to generate features which capture different attention at each block. 

The reason for using residual networks is to have significant influence on increasing the network depth \cite{he2016deep}. This gives the advantage of deeper networks which are easier to train with an increase in accuracy. 
The reason for using attention mechanisms is that they give relative importance to particular sub-parts of the scenarios. For instance, compared to the human-eye mechanism, our brain ignores the information captured during the saccades and accepts the given useful information during fixation. From the given fixations, the brain gives the importance to the segmented useful information. This mechanism is referred to as an attentive behaviour of brain. Similarly, adapting this attention mechanism can help the residual neural network to classify hand gestures with better accuracy than without any attention.

The main contributions of this paper can be summarized as follows: (1) We develop an end to end trainable 3D attention-based deep residual neural network which has stacked attention blocks. The attention features adaptively change with the depth of the network. (2) We provide insights in the number of frames from videos to be input in our network to achieve better results. (3) We perform an evaluation of our network on three different data-sets and illustrate the comparison results. (4) We compare our network with other state of the art networks for hand gesture recognition and also perform ablation study on the number and position of the attention blocks in the baseline network. (5) We also provide suggestions to improve the accuracy of our network by varying the parameters. (6) Finally, we release our PyTorch implementation\footnote{https://github.com/nainadhingra2012/Res3ATN} as an open-source. We also expect that our work will give further advances in hand gesture recognition using attention blocks.

The paper is structured as follows. Related work is discussed in Section 2. The proposed 3D  attention block and residual attention network are described in Section 3. The experiments are elaborately illustrated in Section 4, ablation study is discussed in Section 5 followed by suggestions for improvement in Section 6. Finally, Section 7 concludes our work.


\section{Related Work}
Machine learning methods such as support vector machines (SVMs), Hidden Markov Models (HMMs), decision trees, random forests, and conditional random fields have been implemented in the state of art to classify the hand gestures \cite{molchanov2015hand, dardas2011real, laviola2014introduction, starner1998real, wang2006hidden, trindade2012hand}. Several feature extraction techniques such as hand crafted spatio-temporal features, histogram of gradient features, classical descriptors, etc. were used in the classical machine learning techniques to recognize the gestures \cite{laviola2014introduction, molchanov2015multi}.

Deep Learning techniques have been successfully used in a number of different applications of computer vision \cite{donahue2015long, simonyan2014two} such as object recognition, image segmentation, image classification, image registration, etc. These techniques have outperformed the classical methods by achieving high benchmark results \cite{wang2015action}. Deep learning techniques have also found to be performing well on videos analysis and 3D images analysis for medical purposes. Considering the performance and success of these techniques, we have used them to detect and classify hand gestures on different open source video data sets.

Video analysis and gesture recognition based on deep learning includes mainly \textit{three} techniques based on how the temporal dimensions of the video data are treated \cite{asadi2017survey}.

\textit{First} is a two or more stream data input approach, where two or more streams of data are fed into the 2D conolutional neural network (2D CNNs)  \cite{simonyan2014two,10.1007/978-3-319-46484-8_2, carreira2017quo, sudhakaran2018attention}: 
\begin{itemize}
\item RGB images are encoded as input of one of the two streams.
\item Optical flow is encoded to input as the second stream.
\end{itemize}
Some of the recent work includes feeding more than two streams of the data where the third stream is encoded depth maps, or, extra features \cite{singh2016first, ma2016going, tang2017action}.  

\textit{Second} is the end to end approach, where 3D video data is fed into the network which uses 3D convolutional layers having 3D filters to capture the features from the 3D data along temporal and spatial dimensions. Compare to the 2D CNNs, 3D CNNs are able to extract discriminative features. 2D CNNs have the advantage of using the pre-trained network which is trained on large available 2D datasets \cite{feichtenhofer2016convolutional}. In this paper, we use a 3D CNN  for hand gesture recognition which can take advantage of discriminative features along with temporal and spatial ones.
 
\textit{Third} is the combination of 2D or 3D CNNs and temporal sequence modelling \cite{asadi2017survey}. This combination is then applied to a single frame or stacks of frames using recurrent neural networks or long short term memory (LSTM) as they can capture temporal features using recurrent connections. Several variants such as Hierarchical RNNs (H-RNNs), bidirectional RNNs (B-RNNs), etc and HMMs have been successfully used for temporal modelings \cite{gers2002learning, du2015hierarchical,wu2016deep, veeriah2015differential}. 

There are supervised and unsupervised ways of learning the features from 3D video datasets. Supervised learning uses ground truth data for optimizing  the training process. Unsupervised learning includes extracting the invariant spatio-temporal features from the videos using independent subspace analysis (ISA), autoencoders, or some other variant network \cite{baccouche2012spatio, nandakumar2013multi, marc2007unsupervised}. Convolutional Restricted Boltzmann Machines (RBMs) have also been used for generating feature representations of the video frames \cite{chen2010deep}.

\textbf{Attention Mechanism:} An attention mechanism works based on the functions of the human eye. As we pay attention to a particular region of the total field of view, our brain is trained to interpret the information based on the higher attention area \cite{KingkanOH18} rather than treating the field of view homogeneously. Similarly, attention mechanism have been explored in deep learning specifically for the combination of image analysis and natural language processing applications such as answer selection tasks \cite{tan2015lstm}, image captioning \cite{ xu2015show}, handwriting synthesis \cite{graves2013generating}, machine translation \cite{bahdanau2014neural}, or phoneme recognition \cite{chorowski2014end, chorowski2015attention}. They have also shown to perform well for speech recognition \cite{bahdanau2016end} when supplemented with location-awareness \cite{chorowski2015attention}. 

Attention mechanism are of two different types, namely, soft attention and hard attention. \textit{Soft attention} uses the differentiable function such as softmax function, sigmoid function, etc., and can be trained using backpropagation algorithms \cite{seo2016progressive, wang2017residual, xu2015show, KingkanOH18, sudhakaran2018attention, chen2016attention, jaderberg2015spatial}. Spatial transformer network is an example of soft attention which uses the spatial manipulation of the data \cite{jaderberg2015spatial}. \textit{Hard attention} is based on the non-differentiable stochastic functions such as heaviside step function, switch functions, etc. which have abrupt changes and discontinuities in the functions \cite{gregor2015draw}. 

Our network is based on the soft attention developed for an inclusive feed-forward network in a 2D residual attention network for image classification \cite{wang2017residual}. It uses bottom-up and top-down feed-forward structures which provide soft weights to the features \cite{wang2017residual}.

There has been work performed with 3D residual networks \cite{ahmet, hara2017learning} in the past for gesture recognition which can be scaled to hundreds of layers. To our knowledge, there is no state-of-the-art attention-based 3D residual network used for hand gesture recognition. The 3D residual attention network will have the benefit from residual learning to have deeper networks along with the positives of attention aware features. The attention blocks will help the network to concentrate on the useful region of the video along the spatial as well as the temporal domain. To achieve a better accuracy for recognizing the gestures using the combination of residual network along with attention mechanism, we built our Res3ATN network.


\section{Methodology}
\subsection{3D Convolutional Neural Network}
3D CNNs capture the information from the spatial content as well as the temporal relationship between the different frames of a video. We use 3D CNNs in Res3ATN and in other various networks used for comparison. The C3D network is used as a network to compare with and to evaluate the results of Res3ATN. 

\subsection{Residual Network}
\begin{figure}[t]
\begin{center}
   \includegraphics[width=7cm,height=6cm]{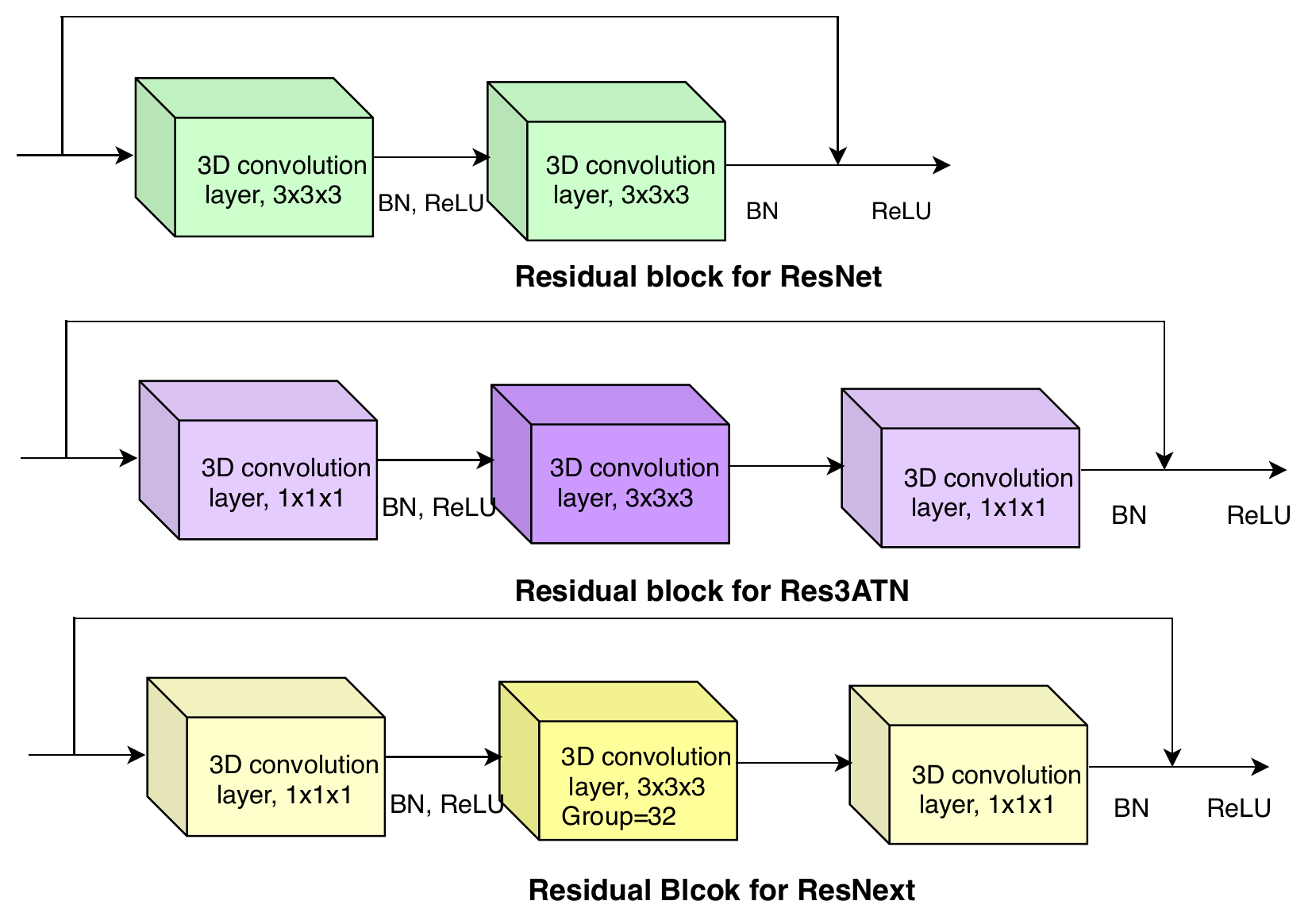}
\end{center}
   \caption{Basic residual blocks used in ResNet-10, Res3ATN and ResNext-101, respectively.}
\label{fig:res_units}
\end{figure}

The residual network have short connections, which are direct connection between two non-consecutive layers \cite{he2016deep}. Since it is easier to optimize the deeper networks using residual connections, it will result in an increased accuracy. Fig. \ref{fig:res_units} shows the residual blocks used for our residual attention network. We use a ResNet-10 and ResNext-101 architecture which is build with the ResNet and ResNext block as shown in Fig. \ref{fig:res_units}.

 For the C3D, ResNet-10 and ResNext-101 models, we use the same architecture as proposed by \cite{ahmet}. But instead of using pre-trained models on Jester datasets, we train our model from scratch on the 3 datasets, i.e., EgoGesture, Jester, and NVIDIA Dynamic Hand Gesture dataset. 

\subsection{3D Residual Attention Network}
Deep residual networks have shown to perform well for very deep networks \cite{he2016identity} with good convergence characteristics. Our network is a 3D residual network with multiple 3D attention blocks. These attention blocks consists of two parts, i.e., trunk and mask layer. We use the trunk layer which consist of residual units \cite{he2016identity, wang2017residual}, but it can be replaced by any other 3D model units. The mask layer consists of the soft attention differentiable function which creates the 3D mask $M(x)$ of the same dimension as the 3D features $T(x)$ generated by the trunk layer. The output $O$ of attention block can be described by equation (\ref{eq:1}). The attention block is differentiable and hence it can be easily trained end to end as we use differentiable soft attention.

\begin{equation}
     O_{i,c,f} = M_{i,c,f} * T_{i,c,f} \label{eq:1}
\end{equation}

\begin{figure*}[htp!]
\begin{center}
   \includegraphics[width=17cm,height=5cm]{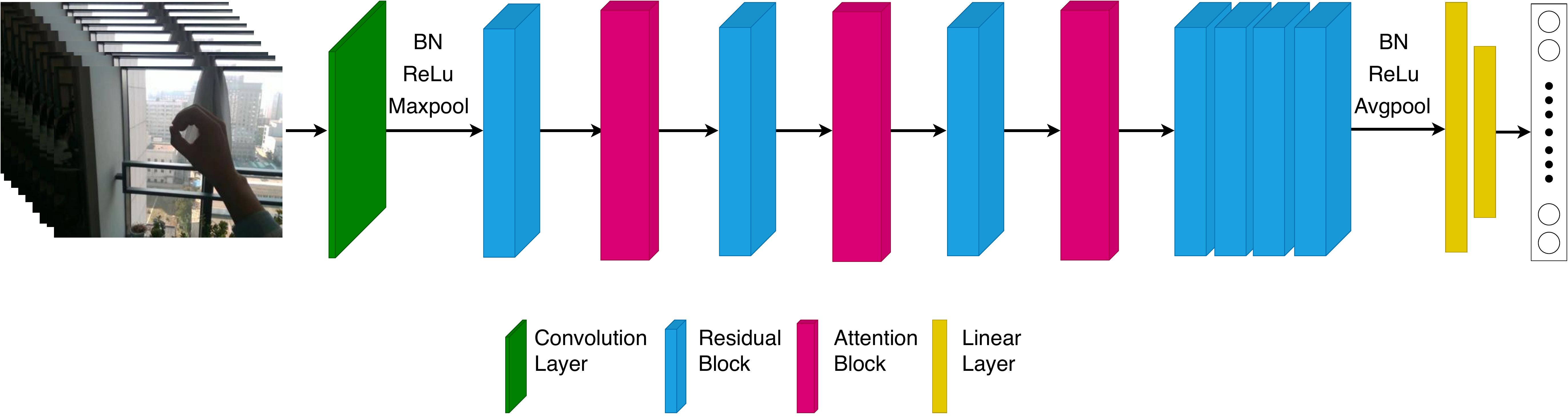}
\end{center}
   \caption{The 3D Residual Attention Network (Res3ATN) used for hand gesture recognition. The input video is fed to the Res3ATN in the form of a stack of 2D images. BN corresponds to batch normalisation, ReLu to rectified linear units and Maxpool to max pooling. The stacks of residual blocks and attention blocks are used to capture the features from the videos followed by two fully connected layers. The last fully connected layer gives outputs equal to the number of classes in each dataset.}
\label{fig:RES_ATN_NETWORK}
\end{figure*}
\begin{figure*}[htp!]
\begin{center}
   \includegraphics[width=17.5cm,height=14.75cm]{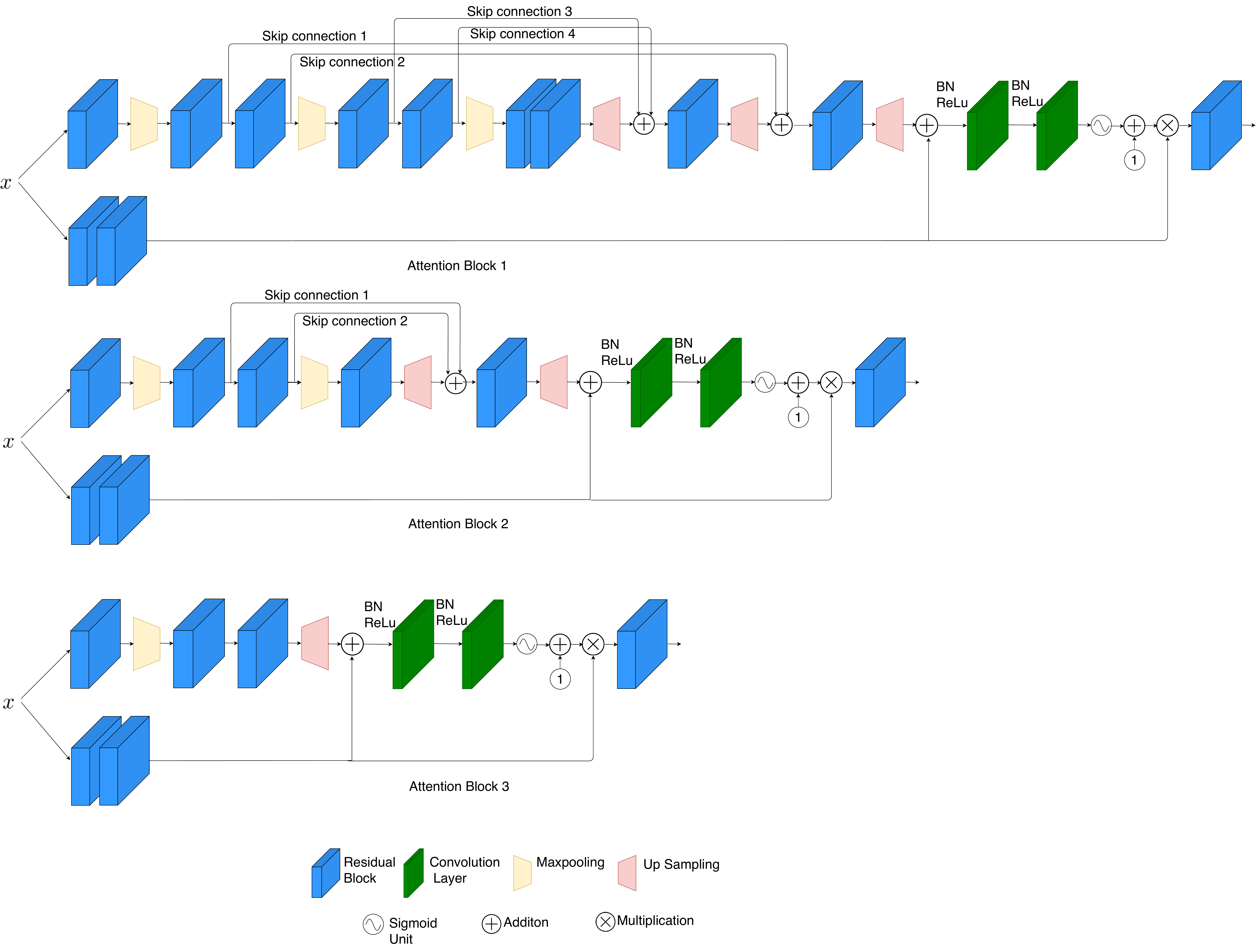}
\end{center}
   \caption{The attention blocks used in the Res3ATN Network. There are multiple residual blocks with the same number of up-sampling as down-sampling blocks. There are four skip connections used in the first attention block. We take skip connections as a parameter, so we reduce the skip connections to two in the second attention block in Res3ATN and zero skip connections in the last attention block in the Res3ATN network. For each max pooling 3D layer, we use a kernel size= 3x3x3 and stride size=2, padding=1. For each residual block, we have the same number of output and input channels as well as 3D-CNN. Each residual block has a kernel size equal to 1x1x1, 3x3x3, 1x1x1, respectively, with stride=1. Two consecutive 3D-CNN layers (green colour box in the figure) in the trunk layer have a kernel size=1x1x1, stride=1x1x1.}
\label{fig:attention}
\end{figure*}

where $i\in \{1,.....,H*W\}$ is the spatial index, $H$ is the height, and $W$ is the width of the 2D frame ,  $c\in \{1,.....,C\}$ is the channel index, and $f\in \{1,.....,F\}$ is the frame index.

The mask layer uses bottom-up and top-down mechanism to get the mask which can manipulate the output features generated by the trunk layer. When the output mask $M$ is multiplied with the $T$, it gives us the weighted features and hence behaves like a feature selector. During the back-propagation, due to its property of differentiability, it updates the gradient. The corresponding mask gradient of the input feature in the soft mask layer is as shown in equation (\ref{eq:2}). If the trunk features $T$ are not correct, mask can prevent \cite{wang2017residual}  $T$ features to update the parameters as there is a multiplication factor of the mask $M$ with partial derivative of $T$ as shown in equation (\ref{eq:2}).

\begin{equation}
\frac{\partial M(x, \theta_m)T(x,\theta_t)}{\partial \theta_t} = M(x, \theta_m)\frac{\partial T(x,\theta_t)}{\partial \theta_t}\label{eq:2}
\end{equation}

where $x$ is the input, $\theta_t$ are trunk layer parameters and  $\theta_m$ are mask layer parameters. We stack number of attention blocks to filter out different useful features. For the task of hand gesture detection, the network should detect where it should concentrate its attention and then recognize the hand gesture class such as moving hand up, down, right or left, etc. So, the attention block helps the residual attention network to identify the region where the gesture is performed in each frame of a video. The use of multiple attention blocks makes the network more robust as it is able to capture different types of attention focusing at different types of features at each attention block. Since the gesture detection in videos is a difficult task considering the number of features the network has to learn for a single video as compared to a single image, the multiple attention network alleviates this problem by learning multiple masks.

Motivated by the techniques used by \cite{wang2017residual}, to keep the identity function of the residual network, instead of using equation (\ref{eq:1}) as it is, we add "1" to the generated mask $M$. This leads to revive the functionality of the residual network and at the same time it does not hamper the performance due to the dot product between trunk output $T$ with the zeros from the mask $M$. So, it modifies equation (\ref{eq:1}) to equation (\ref{eq:3}), which is similarly done by \cite{wang2017residual} for the 2D images classification.

\begin{equation}
     O_{i,c,f} = (1+M_{i,c,f}) * T_{i,c,f} \label{eq:3}
\end{equation}

Similar to the bottom-up and top-down structure in the famous U-Net \cite{ronneberger2015u, cciccek20163d, dong2017automatic} network used for segmentation tasks, we use the same approach in our attention block which is used for achieving a good mask that could act as a filter to the trunk layer features. The output of the mask layer has the same dimensions as the output from the trunk layer. So, the number of 3D max-pool layers \cite{szegedy2015going}, which are used for down-sampling, are the same as for the 3D interpolation layers. As suggested in \cite{wang2017residual}, we used skip connections to connect the bottom-up and top-down structures to get the feature information from various scale levels.

Regarding the 2D results from \cite{wang2017residual}, we have mixed attention, i.e., spatial, channel, and frame attention by using a Sigmoid function \cite{han1995influence} for each point in the 3D data $x$, given by:

\begin{equation}
f_(x_{i,c,f}) = \frac{1}{1+ exp(-x_{i,c,f})}\\
\end{equation}
 
 where $i\in \{1,.....,H*W\}$ is the spatial index, $c\in \{1,.....,C\}$ is the channel index, and $f\in \{1,.....,F\}$ is the frame index.

The 3D residual attention network adaptively changes attention as the feature changes. Each attention block learns and captures different types of features which avoids errors or wrong focus of attention. Moreover, a wrong attention predicted by one block can be corrected by the other attention blocks. Thus, multiple blocks make the network quite robust on the attention prediction. 

\begin{figure*}[htp!]
\begin{center}
   \includegraphics[width=16cm,height=10cm]{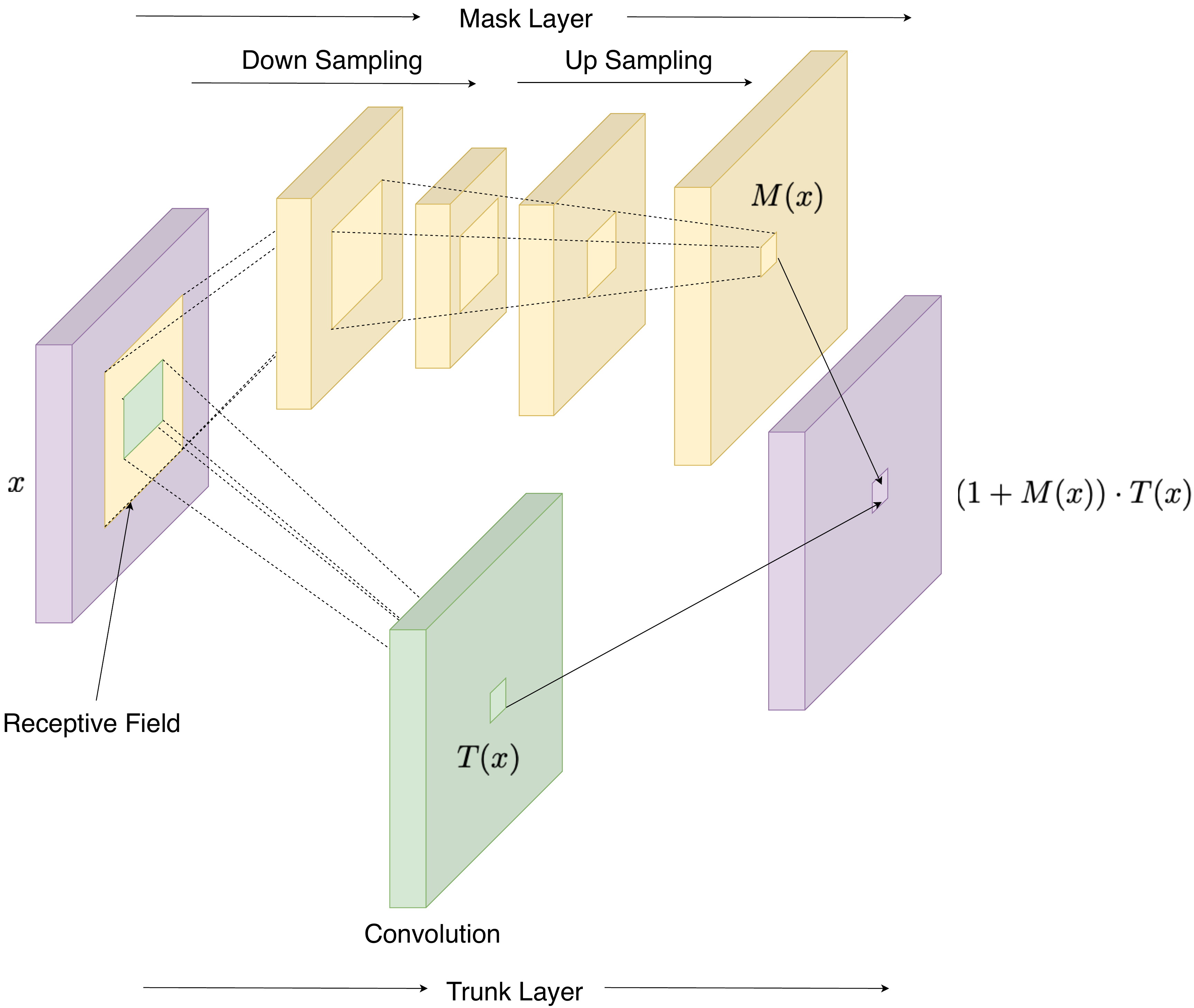}
\end{center}
   \caption{Receptive field through the 3D attention block is similar to the receptive field for 2D attention block in \cite{wang2017residual}. The receptive field decreases from input to the output mask in the mask layer and also similarly decreases for the trunk layer. The output of the trunk and mask layer have the same dimensions.}
\label{fig:onecol}
\end{figure*}

Our architecture of Res3ATN (see Fig. \ref{fig:RES_ATN_NETWORK}) is similar to \cite{wang2017residual}, but we added one extra fully connected layer in their 2D residual attention network. We implemented our network for 3D application of hand gesture in videos so we used 3D-CNN instead of 2D-CNN, 3D Max-pool instead of 2D Max-pool and likewise 3D interpolation  and 3D batch-normalisation. Also, the number of filters, kernel size and strides used for each layer are different from the ones used by them as shown in Table \ref{tab:Features}. The 3 different attention blocks used in our Res3ATN are shown in Fig. \ref{fig:attention}. We reduce the number of residual blocks, skip connections, upsampling and downsampling blocks in attention 2 block as compared to attention 1 block and further reduce for attention 3 block. 

\begin{table}\small
\setlength{\abovecaptionskip}{0pt}
\setlength{\belowcaptionskip}{-5pt}
\begin{center}
\begin{tabular}{lr} \hline

\textbf{Layer}&  \textbf{Res3ATN} \textbf{Network}  \\

\hline
Conv &   3x3x3, 64, stride=1\\
\hline
 MaxPool3D  & 3x3x3, stride=2\\
 \hline
 Residual Block & 1x1x1, 32, stride=1\\
                 & 3x3x3, 32, stride=2\\
                  & 1x1x1,128, stride=1\\
\hline
Attention Block 1 & 128\\
\hline
Residual Block & 1x1x1, 64, stride=1\\
                & 3x3x3, 64, stride=2\\
                & 1x1x1, 256, stride=1\\
 \hline
Attention Block 2 & 256\\
\hline
Residual Block & 1x1x1, 128, stride=1\\
                & 3x3x3, 128, stride=2\\
                & 1x1x1, 512, stride=1\\
 \hline
Attention Block 3 & 512\\  
 \hline
Residual Block & 1x1x1, 256, stride=1\\
                & 3x3x3, 256, stride=2\\
                & 1x1x1, 1028, stride=1\\
 \hline
 Residual Block & 1x1x1, 256, stride=1\\
                & 3x3x3, 256, stride=1\\
                & 1x1x1, 1028, stride=1\\
 \hline
 Residual Block & 1x1x1, 256, stride=1\\
                & 3x3x3, 256, stride=1\\
                & 1x1x1, 1028, stride=1\\
 \hline
 Residual Block & 1x1x1, 512, stride=1\\
                & 3x3x3, 512, stride=1\\
                & 1x1x1, 2048, stride=1\\
 \hline
 Average3D Pool & 2x2x2, stride=2\\
 \hline
 Fully Connected layer & 512\\
 \hline
 Fully connected layer & output classes\\
\hline
\end{tabular}
\end{center}
\caption{Res3ATN Network configuration.}
\label{tab:Features}
\end{table}


\section{Experiments}
In this section, we will elaborately explain our experiments and their results on our 3D residual attention network for hand gesture recognition in different video datasets. The performance of the Res3ATN network is tested on three open sourced datasets: EgoGesture  \cite{zhang2018egogesture}, Jester \cite{Jester2019}, and NVIDIA Dynamic Hand Gesture dataset \cite{molchanov2016online}. We compare and evaluate the performance of the Res3ATN with three other networks, i.e., C3D, ResNet-10, ResNext-101.

\subsection{Training Details}
To evaluate all the compared networks fairly, we used the same experimental conditions for all of them. We used a learning rate of 0.01, which was kept constant throughout the training. Further, we used Nesterov Stochastic Gradient Descent (SGD) optimizer with a momentum of 0.9 and weight decay of 0.001. We trained the networks for 30 epochs. Next, we used data augmentation techniques for training our networks similar to ones used by \cite{ahmet}. We randomly crop the image to a size of 112 x 112 px. Each image is scaled with either $1, 1/2^{1/4}, 1/2^{3/4}$ or $1/2$  randomly and then we apply spatial elastic displacement having $\sigma=2$ and $\alpha=1$ after cropping the image. We also select the set of defined numbers of frames from the video containing the gestures randomly. The normalization of images is performed to have the pixel values between 0-1 scale value. We use the same training details for all the networks which are used for comparison and evaluation of Res3ATN network.

\subsection{Evaluation using EgoGesture Datasets}

\begin{table}\small
\setlength{\abovecaptionskip}{0pt}
\setlength{\belowcaptionskip}{-5pt}
\begin{center}
\begin{tabular}{cccc} \hline
& EgoGesture & {Jester} & {NVIDIA dataset}  \\
\hline
Classes & 83 &26 &25\\
 Total& 2,081  & 148,092  & 1532\\

 Train &  1239 & 118,562 & 1050\\

 Valid&   411& 14,787 & -\\
 Test& 431 &14,743 & 482\\
\hline
\end{tabular}
\end{center}
\caption{Overview of EgoGesture, Jester and NVIDIA dynamic hand gesture dataset.}
\label{tab:activation_exp}
\end{table}

\textit{EgoGesture dataset} has 83 classes depicting 83 different hand gestures in videos which are captured in a mixture of diverse indoor and outdoor scenes with 50 different subjects. These videos show interaction of wearable devices with hands. A total of 6 scenes was used, 4 scenes were outdoor and 2 scenes were indoor. The total dataset contains  2,081 RGB-D videos which is divided into training, validation, and testing sets with the 3:1:1 ratio randomly based on the subjects. 1,239 training videos have 14,416 gesture samples, 411 validation videos have 4,768 gesture samples and 431 testing videos have 4'977 gesture samples. Some of gesture examples in the videos are: scroll hand downward, number 0-9, applaud, walk, move hand towards right, scroll fingers towards left, etc.  

\begin{table}\small
\begin{center}
\begin{tabular}{ccccc}
\hline
{Model} & {Frames} & {Modality} & {Top-1 acc} & {Top-5 acc}  \\
\hline

C3D   & 32 & Depth & 92.11 &  98.20 \\
ResNet-10  & 32 & Depth& 91.56   & 97.84\\
ResNext-101  & 32& Depth&  92.24 & 98.32 \\
Res3ATN-56 & 32 & Depth & 93.63 & 98.65 \\
\hline

\end{tabular}
\end{center}
\caption{Comparison of the Res3ATN model with different network configurations using the EgoGesture dataset.}
\label{tab:egoresults}
\end{table}

\begin{table}\small
\begin{center}
\begin{tabular}{cccc}
\hline

{No. of frames}  & {8} & {16 }& {32} \\
\hline
 Top-1 acc & 79.28 &  88.38 & 93.63 \\

\hline
\end{tabular}
\end{center}
\caption{Comparison of performance of Res3ATN with different number of input frames.}
\label{tab:frames}
\end{table}

EgoGesture is a large egocentric dataset being publicly available. We used depth modality images for our training and evaluation of all the four networks because \cite{ahmet} stated that they achieved better results with EgoGesture depth images than EgoGesture RGB images. The reason for this is that the depth sensors concentrate on the hand motion and neglect the background motion. The hand gesture detection results by all the four networks is described in Table \ref{tab:egoresults}. It can be seen that Res3ATN outperforms all the 3 other networks in our experiments. 
We also investigated the number of input frames for the Res3ATN. The detailed results of input frames is shown in Table \ref{tab:frames}. It shows that the 32 frames have better results than 8 or 16 input frames. We use 32 frames as input to all the networks.

\subsection{Evaluation using Jester Datasets}

\textit{Jester dataset} is a large set of videos with hand gestures performed by humans in front of the webcam or a laptop camera. It is a crowd-sourced data base with large number of people contributing to it. It is publicly available for research purposes. It has 27 classes with total of 148,092 RGB videos. from which 118,562 are used for training, 14,787 videos for validation and 14,743 videos for testing. Some of the examples of the classes are: Pulling Hand In, Pushing Two Fingers Away, Shaking Hand, Sliding Two Fingers Up, Swiping Down, etc.

We investigated the performance of C3D, ResNet-10, ResNext-101 and Res3ATN using the Jester dataset. The Jester dataset is the largest of all the three datasets. We use the RGB modality videos of this dataset. We noted that ResNext-101 performed better than the ResNet-10 and C3D because of the larger network depth in ResNext-101, which is able to learn more descriptive features for the large dataset. Our network, Res3ATN performs better than the other three networks as shown in Table \ref{tab:Jester}.

\begin{table}\small
\begin{center}
\begin{tabular}{ccccc}
\hline
{Model} & {Frames} & {Modality} & {Top-1 acc} & {Top-5 acc}  \\
\hline

C3D   & 32 & RGB & 75.20  & 85.16 \\

ResNet-10  & 32 & RGB &  78.22  & 87.03\\
ResNext-101  & 32& RGB &  82.24 & 89.01 \\
Res3ATN & 32 & RGB & 84.56 & 91.95 \\
\hline

\end{tabular}
\end{center}
\caption{Comparison of Res3ATN with different models for the Jester dataset.}
\label{tab:Jester}
\end{table}

\subsection{Evaluation using NVIDIA Dynamic Hand Gesture Dataset}

\textit{NVIDIA Dynamic Hand Gestures dataset} has 25 hand gesture types. Each video is recorded by multiple sensors from different viewpoints. A total of 1532 dynamic hand gesture videos are captured by 20 subjects in an indoor environment, varying the intensity of the light using a car simulator with different lightning conditions. The dataset contains 70 \% training videos, i.e., 1050 videos and 30 \% testing videos, i.e., 482 videos. The split is created randomly on the data based on the subjects. Some of the gestures are: showing the two or three fingers, shaking the hand, pushing the hand up, rotating two fingers clockwise or counterclockwise, etc.

\begin{table}\small
\begin{center}
\begin{tabular}{ccccc}
\hline
{Model} & {Frames} & {Modality}  & {Top-1 acc} & {Top-5 acc}  \\
\hline

C3D   & 32 & RGB & 53.94 & 71.16 \\
ResNet-10  & 32 & RGB &  56.74 & 76.03\\
ResNext-101  & 32& RGB &  51.24 & 64.00 \\
Res3ATN & 32 & RGB & 62.65 & 81.95 \\
\hline

\end{tabular}
\end{center}
\caption{Comparison of Res3ATN with different network configurations for NVIDIA Dynamic Hand Gesture dataset.}
\label{tab:NVIDIA}
\end{table}

We investigated the performance of Res3ATN network using NVIDIA Dynamic Hand Gesture dataset, which is a very small dataset compared to the EgoGesture and Jester datasets. Because of the small size, the performance of all four networks is low as compared to the performance of the other two datasets. The detailed results are shown in Table \ref{tab:NVIDIA}. However, the evaluation accuracy can be increased by using pre-trained networks on large datasets and then train on the NVIDIA Dynamic Hand Gesture dataset.  


\section{Ablation study on multiple attention blocks}
In this section, we compare Res3ATN with its baseline network which has same structure as Res3ATN but has zero attention blocks. We evaluate networks having different numbers of attention blocks and also investigate results with different positions of attention blocks for the same network.  The results are shown in Table \ref{tab:attention_blocks_n}. It is evident that the Res3ATN performs better than the networks having 1 or 2 blocks. The reason for the better performance of Res3ATN is that multiple attention blocks tend to capture the correct corresponding attention even if one of the attention blocks is capturing wrong features. Hence, results into making Res3ATN robust for the prediction of attention.

We also compared the location of the attention block in case of 1 attention block in the baseline network. Table \ref{tab:attention_blocks_1} indicates that the location of second attention block of the network in Fig. \ref{fig:attention} performs better than the first and the third attention, when used only one attention block at a time.

While evaluating the networks having 2 attention blocks at a time, we analyzed that the baseline network having the first and the third attention block performs better than the one having the first and the second attention block, and the network having the second and the third attention block as described in Table \ref{tab:attention_blocks_2}.

\begin{table}\small
\begin{center}
\begin{tabular}{cccc}
\hline
{Model} & {Attention} & {Top-1 acc} & {Top-5 acc}  \\
{ } & {blocks} & { } & { }  \\
\hline

Baseline with 0 ATN  &  0 & 51.45 & 64.10 \\
Baseline with 1 ATN  & 1 & 52.38   & 67.73 \\
Baseline with 2 ATN  & 2 &  52.70   & 68.26\\
Res3ATN &  3 & 62.65 & 81.95 \\
\hline

\end{tabular}
\end{center}
\setlength{\belowcaptionskip}{-40pt}

\caption{Comparison of networks having different number of attention blocks. The baseline network is the Res3ATN without any of the attention blocks, Baseline with 1 ATN is the baseline network with 1 attention block. Baseline with 2 ATN is the baseline network with 2 attention blocks.}
\label{tab:attention_blocks_n}
\end{table}

\begin{table}\small
\begin{center}
\begin{tabular}{cccc}
\hline
{Model} & {Attention blocks} &  {Top-1 acc} & {Top-5 acc}  \\
\hline

$1_0$   &  1 & 47.30  & 62.03 \\
$1_1$  & 1 &52.38  &  67.73 \\
$1_2$ & 1 &  51.86 & 65.87 \\
\hline

\end{tabular}
\end{center}
\caption{Comparison of networks having 1 attention block at different positions in a baseline network. The model name $1_0$ refers to the baseline network the having first attention block of the Res3ATN. $1_1$ refers to the baseline network having the second attention block of the Res3ATN. Similarly, $1_2$ refers to the network having the third attention block of the Res3ATN.}
\label{tab:attention_blocks_1}
\end{table}

\begin{table}\small
\begin{center}
\begin{tabular}{cccc}
\hline
{Model} & {Attention blocks} &  {Top-1 acc} & {Top-5 acc}  \\
\hline

$2_0$   &  2 & 50.83 & 66.49 \\
$2_1$  & 2 & 52.70   & 68.26\\
$2_2$ & 2 &  51.66 & 65.45 \\
\hline

\end{tabular}
\end{center}
\caption{Comparison of networks with 2 attention blocks at different positions in a baseline network. The model name $2_0$ refers to the baseline network having the first two attention blocks of the Res3ATN. $2_1$ refers to the baseline network having the first and third attention blocks of the Res3ATN, while $2_2$ refers to the network having the second and third attention blocks of the Res3ATN.}
\label{tab:attention_blocks_2}
\end{table}

\section{Further Improvements}
Further improvements are possible by using larger input frames than just 112 x 112 px size, since an increase of the spatial resolution and the number of frames can improve the classification accuracy \cite{varol2017long, hara2017learning}. Due to GPU limitations and considering the size of the Res3ATN, we only used 112 x 112 px size. With the use of batch-normalization in the networks, it is important to train models with large batch-sizes \cite{carreira2017quo,hara2017learning}. We use batch-normalization in our network architecture, so it is better to have larger batch-size for training. However, we only used a small batch-size of 6 due to GPU limitations. Since activity recognition is a very related task to hand gesture recognition, using pre-trained models trained on activity recognition datasets such as activity-net or UCF-101 might help to increase the accuracy of the hand gesture recognition network.


\section{Conclusion}
In this paper, we have successfully developed Res3ATN, i.e., 3D attention based residual network for hand gesture recognition. We validated the performance of Res3ATN on three  publicly available hand gesture datasets. The proposed techniques performs better than the three other widely used networks, i.e., C3D, ResNet-10, ResNext-101 for 3D activity and gesture recognition. We investigated the number of frames which should be input to the Res3ATN. We inspected the number of attention blocks and the position of the attention blocks to be used in the network. Our analysis shows that the stacked multiple soft attention blocks help the network to recognize the hand gestures with better accuracy. For future work, we will use pre-trained networks which are trained on different activity recognition datasets and evaluate the performance of Res3ATN. 

\section*{Acknowledgements}
This work has been supported by the Swiss National Science Foundation (SNF) under the grant no. 200021E\_177542 / 1.  It is part of a joint project between TU Darmstadt, ETH Zurich, and JKU Linz with the respective funding organizations DFG (German Research Foundation), SNF (Swiss National Science Foundation) and FWF (Austrian Science Fund).

{\small
\bibliographystyle{ieee}
\bibliography{egbib}
}

\end{document}